\documentclass[sigconf,natbib=true]{acmart}
\AtBeginDocument{%
  }

\setcopyright{acmlicensed}
\copyrightyear{2026}
\acmYear{2026}
\acmDOI{XXXXXXX.XXXXXXX}
\acmConference[CIKM 2026]{The 35th ACM International Conference on Information and Knowledge Management}{November 7--11, 2026}{Rome, Italy}
\acmISBN{978-1-4503-XXXX-X/2026/11}

\usepackage{times}
\usepackage{soul}
\usepackage{url}
\usepackage[utf8]{inputenc}
\usepackage{graphicx}
\usepackage{amsmath}

\usepackage{amsthm}
\usepackage{booktabs}
\usepackage[linesnumbered,ruled,vlined]{algorithm2e}
\usepackage{amssymb}
\usepackage{makecell}
\usepackage{subfigure}

\newcommand{\stitle}[1]{\vspace{0.5em}\noindent\textbf{#1}}

\begin{document}

\title{Efficient Coreset Selection via K-Nearest Neighbor Graphs}


\author{Yingfan Liu}
\affiliation{%
  \institution{Xidian University}
  \city{Xi'an}
  \country{China}
}
\email{liuyingfan@xidian.edu.cn}

\author{Leiyu Zhang}
\affiliation{%
  \institution{Hangzhou Institute, Xidian University}
  \city{Hangzhou}
  \country{China}
}
\email{23031212287@stu.xidian.edu.cn}

\author{Jiadong Xie}
\authornote{Jiadong Xie and Mingzhe Wang are the corresponding authors.}
\affiliation{%
  \institution{The Chinese University of Hong Kong}
  \city{Hong Kong SAR}
  \country{China}
}
\email{jdxie@se.cuhk.edu.hk}

\author{Mingzhe Wang}
\authornotemark[1]
\affiliation{%
  \institution{Xidian University}
  \city{Xi'an}
  \country{China}
}
\email{wangmingzhe@xidian.edu.cn}

\author{Jeffrey Xu Yu}
\affiliation{%
  \institution{The Hong Kong University of Science and Technology (Guangzhou)}
  \city{Guangzhou}
  \country{China}
}
\email{jeffreyxuyu@hkust-gz.edu.cn}

\author{Jiangtao Cui}
\affiliation{%
  \institution{Xi'an University of Posts and Telecommunications}
  \city{Xi'an}
  \country{China}
}
\affiliation{%
  \institution{Xidian University}
  \city{Xi'an}
  \country{China}
}
\email{cuijt@xidian.edu.cn}

\renewcommand{\shortauthors}{Liu et al.}

\begin{abstract}
Coreset selection reduces the cost of model training by replacing a large training set with a small representative subset. Existing gradient-approximation coreset methods such as CRAIG and cluster-based variants can preserve model accuracy. Still, their selection stages often rely on dense pairwise distances or large item-cluster bound matrices, leading to high time and memory costs on large datasets. This paper proposes KNNG-CS, a lightweight coreset selection method based on a $K$-nearest neighbor graph. KNNG-CS exploits local neighborhood structures to estimate the importance of each data item and greedily selects representative nodes without maintaining a quadratic distance matrix. 
The method requires only linear storage in the number of edges.
Experiments on four real-world datasets show that KNNG-CS achieves accuracy comparable to representative gradient-approximation coreset methods, while reducing selection time by $2.3\times$-$41.2\times$ and peak memory to $0.3\%$-$7.5\%$ of the baselines.
\end{abstract}




\maketitle

\section{Introduction}

\label{sec:intro}

Coreset selection is a widely used technique for reducing the cost of training machine learning models. 
Given a large training set, it selects a small subset that preserves the training utility of the full data, so that the model can be trained with lower computation and memory consumption~\cite{jure,cluster,k-means-clustering,svm-clustering,wei2015submodularity}. 
This is especially useful when data selection can be performed before model training, where the selected subset can be reused across multiple training runs or hyperparameter settings.

In this paper, we focus on gradient-approximation (GA) based coreset selection~\cite{jure,cluster}. 
Given a training dataset $T=(t_1,t_2,\cdots,t_n)$, where each item $t_i$ contains a feature vector $x_i\in\mathbb{R}^d$ and a label $y_i$, GA-based methods select a subset $C\subseteq T$ with $|C|\le M$ to approximate the gradients of all training items. 
Following prior studies~\cite{allen2016exploiting,hofmann2015variance,jure,cluster}, the GA error can be upper bounded by feature-space distances, and the problem can be reduced to minimizing the following feature-space representative error:
\begin{equation}
C^{*}=\arg\min_{C\subseteq T,\ |C|\le M}
\sum_{i=1}^{n}\min_{t_j\in C}\|x_i-x_j\|.
\label{eq:5}
\end{equation}
Here, each training item $t_i$ is represented by its feature vector $x_i$, and the objective measures the total distance from every item to its nearest selected representative in $C$. Since Eq.~\ref{eq:5} is independent of model parameters, the coreset can be selected without prior training.

Since solving Eq.~\ref{eq:5} exactly is NP-hard~\cite{jure}, existing GA-based methods typically rely on greedy selection.
Given a current coreset $C$, the greedy framework evaluates the benefit of adding each candidate item and repeatedly selects the one with the largest benefit until $M$ items are selected. 
CRAIG~\cite{jure} follows this framework and relies on pairwise feature distances to compute candidate benefits. 
To avoid repeated distance computations, it pre-computes a distance matrix, requiring $O(n^2d)$ time and $O(n^2)$ space. 
Although CRAIG can use a divide-and-conquer strategy to reduce memory consumption, it still suffers from high computational and storage costs on large datasets. 
FastCore~\cite{cluster} improves efficiency by sampling candidate items and replacing item-item distances with item-cluster upper bounds. 
However, it usually requires a large number of clusters, e.g., hundreds to tens of thousands, to balance accuracy and efficiency, which still incurs considerable computation and memory overhead.

Our key observation is that feature-space representativeness is inherently local. 
A sample that appears as a close neighbor of many other samples is likely to represent a dense local region with small approximation error, while isolated samples are less likely to serve as useful representatives for many points. 
However, existing greedy methods mainly exploit this information through dense distance structures or large cluster-bound matrices, rather than directly modeling local neighborhood relations in a sparse form.

Motivated by this observation, we propose \textsc{KNNG-CS}, a lightweight coreset selection method based on a $K$-nearest neighbor graph (KNNG)~\cite{KGraph}. 
\textsc{KNNG-CS} first builds an approximate KNNG over the feature vectors, where each node stores $K$ local outgoing neighbors, capturing local similarity relationships in a sparse structure.
It then estimates node importance by aggregating distance-aware incoming-neighbor votes: a node receives a high score if many samples regard it as a close local neighbor. 
Finally, \textsc{KNNG-CS} performs greedy selection by repeatedly selecting high-score nodes and removing locally covered nodes from further consideration. 
This design avoids the quadratic distance matrix used by CRAIG and the large item-cluster bound matrix used by FastCore, requiring only $O(nK)$ graph storage with a small value of $K$.

Our main contributions are summarized as follows.
\begin{itemize}
    \item We propose \textsc{KNNG-CS}, a scalable GA-based coreset selection method that exploits KNNG to avoid expensive dense distance structures.
    \item We design a distance-aware incoming-neighbor scoring function that estimates local representativeness from a KNNG without dense pairwise distance storage.
    \item We conduct experiments on real-world datasets, showing that \textsc{KNNG-CS} achieves $2.3$--$41.2\times$ speedups and uses only $0.3\%$--$7.5\%$ peak memory compared with state-of-the-art methods, while preserving coreset quality.
\end{itemize}

\section{The KNNG-CS Method}
\label{sec:ours}


In this section, we first briefly introduce KNNG and then present our coreset selection method based on KNNG. 


\begin{figure}[t]
    \centering
    \subfigure[Illustration of KNNG]{
        \includegraphics[width=0.35\textwidth]{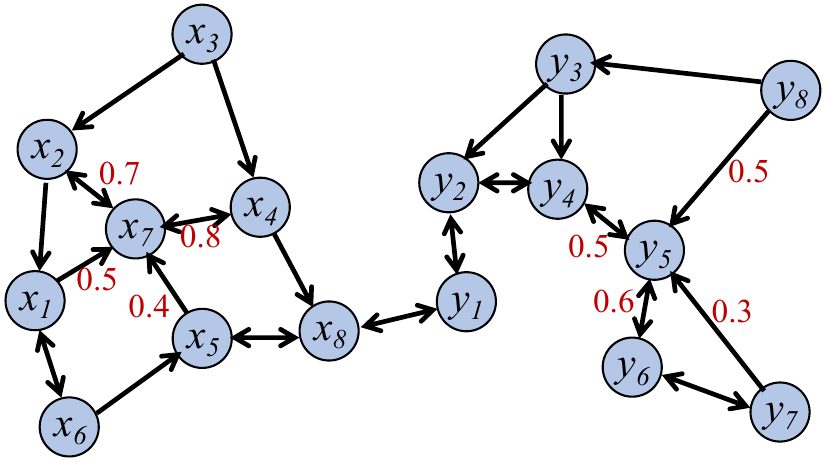}
        \label{fig:knng}
    }
    \hspace{10mm}
    \subfigure[Illustration of node removal]{
        \includegraphics[width=0.35\textwidth]{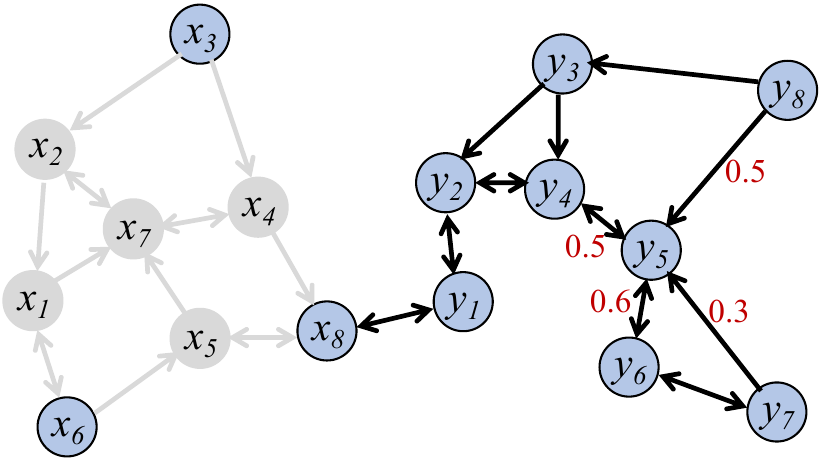}
        \label{fig:node-removal}
    }
    \caption{The illustration of our method.}
    \label{fig:knng-cs}
\end{figure}

\subsection{K-Nearest Neighbor Graph}
\label{ssec:knng}

K-nearest neighbor graph (KNNG) is a graph structure that captures similarity relationships among vectors. For simplicity, we denote $X=\{x_1, x_2, \ldots, x_n\}$, where $x_i \in \mathbb{R}^d$ is the feature vector of $t_i \in T$. Given $X$ and a positive integer $K$, the KNNG $G=(V, E)$ for $X$, where $V$ is the vertex set and $E$ the edge set, treats each vector $x_i \in X$ as a graph node and then builds directed edges between $x_i$ and its $K$-nearest neighbor w.r.t. $X \setminus \{x_i\}$. 
For simplicity, we use the feature vector $x_i$ to represent the corresponding graph node in KNNG. 
We denote $N_G^{out}(x_i) = \{x_j | (x_i, x_j) \in E\}$ as the out-neighbor set of $x_i$ and $N_G^{in}(x_i) = \{x_j | (x_j, x_i) \in E\}$ as its in-neighbor set. 
We illustrate a KNNG built on a set of 16 vectors with $K=2$ in Figure~\ref{fig:knng}. Let us consider $x_7$. We have that $N_G^{out}(x_7) = \{x_2, x_4\}$ and $N_G^{in}(x_7) = \{x_1, x_2, x_4, x_5\}$. 

Due to the ``curse of dimensionality'', the construction of an exact KNNG costs $O(n^2d)$, which is prohibitive, especially on large high-dimensional datasets. Fortunately, there exists a bulk of methods~\cite{Overlap,KGraph,LargeVis2016,MyKNNGCsurvey} that efficiently build approximate KNNGs while slightly sacrificing the accuracy. The time complexity of building an approximate KNNG is $O(n^{1.14}d)$~\cite{Overlap}, while the space complexity of KNNG is $O(nK)$. 
Notably, $K$ is quite small in practice, which is set as 10 in our experiments. 
In this work, we employ the state-of-the-art method~\cite{MyKNNGCsurvey}, i.e.,  KGraph~\cite{KGraph,kgraph-code}, to build the KNNG. Thus, we can efficiently build a KNNG for a large dataset while requiring much less space to store it than the distance matrix in the greedy framework.

\subsection{KNNG-Based Coreset Selection}

Unlike existing works where each item is considered with an equal probability of appearing in the finally generated $C$, we believe that the items should be treated unequally and thus estimate the importance for each graph node (i.e., each item) w.r.t. the goal in Eq.~\ref{eq:5}. With the node importance based on KNNG, we design a greedy method to efficiently select the $M$ most important nodes/items to form $C$. 


\stitle{Local Representativeness:}
According to Eq.~\ref{eq:5}, coreset selection is transferred to select $M$ representative vectors from the set $X$ of feature vectors, where each vector $x_i \in X$ should be assigned to the closest one in $C$ to it, i.e., the 1-nearest neighbor (1-NN) of $x_i$ in the coreset $C$.
However, the set of all 1-NN of the vectors in $X$ may contain more than $M$ vectors, especially when $M$ is small enough. As a result, to meet the limit $M$, a part of vectors in $X$ would choose their 2nd closest neighbor or even farther one which may appear in $C$.
Since KNNG records the KNN of each vector, we can use it to efficiently find the coreset with high quality. 

Intuitively, the node with a larger in-degree has a higher probability than one with a smaller in-degree. This is because the in-degree $|N_G^{in}(u)|$ indicates that $u$ is among the KNN of $|N_G^{in}(u)|$ other nodes. If we select $u$ in the coreset, at least $|N_G^{in}(u)|$ nodes will choose $u$ as their representative vector. However, only the in-degree is insufficient in some cases. Consider that two nodes $u, v\in V$ and $|N_G^{in}(u)| = |N_G^{in}(v)|$, where the in-degree fails to help us distinguish which one is more important for our goal in Eq.~\ref{eq:5}. 
To address this issue, we have to take the distance information of each edge in KNNG into account. In this way, for $u$ and $v$ with equal in-degree, the one with shorter distances is more important for our goal. As shown in Figure~\ref{fig:knng}, we can see that $|N_G^{in}(x_7)| = |N_G^{in}(y_5)| = 4$. Besides, the in-neighbors of $x_7$ have overall shorter distances than those of $y_5$. Hence, $x_7$ is more important than $y_5$. 



\stitle{Distance-Aware Node Scoring:} 
As discussed above, our node importance estimator for each $u\in V$ considers two factors, (1) its in-degree $|N_G^{in}(u)|$ and (2) the Euclidean distances between $u$ and the members in $N_G^{in}(u)$. 
Here, we employ the softmax function to convert the distance of an edge in $G$ into a probability. To be specific, we define the probability of the edge from $v_i$ to $v_j$ as follows. 
\begin{equation}
p_{ij}
= \frac{\exp(-||v_i, v_j||/T)}{\sum_{k \in N^{{out}}_G(v_i)} \exp(-||v_i, v_k||/T)},
\label{eq:9}
\end{equation}
where $T$ is the temperature parameter~\cite{gao2017properties}.
According to Eq.~\ref{eq:9}, the value $p_{ij}$ can be interpreted as the normalized closeness 
weight that $v_i$ assigns to its neighbor $v_j$ among its $K$ outgoing neighbors.
With the $p_{ij}$ values, we define the node importance of each node $v_i$ as follows:
\begin{equation}
    P_i = \sum_{v_j \in N_G^{in}(v_i)} p_{ji}.
\label{eq:np}
\end{equation}
Here, we take the probability of each edge as a weight and compute the node importance of a node $v_i$ as the sum of the weights of all its in-neighbors. In this way, the estimator of the node importance considers both the in-degree and the edge distances simultaneously. 
Notably, we do not directly use the Euclidean distance as the weight of each edge and thus estimate the node importance by the sum of the in-neighbors. This is because the distance values have a larger range. By the softmax function, we can limit the weight of each node in the range $(0, 1)$ and the sum of the out-neighbor weights for each node is 1.

\begin{algorithm}[t]
\caption{Coreset Selection based on KNNG}
\KwIn{Training Set $T$, neighbor size $K$ and coreset size $M$.}
\KwOut{Coreset $C$, $\Lambda = \{\lambda_j\}$, where $|C| = |\Lambda| = M$.}
$C \leftarrow \emptyset$\;
Compute the KNNG $G=(V, E)$ over $X$ via KGraph~\cite{KGraph}\;
\For{each node $v_i \in V$}{
    \For{each $v_j \in N_G^{out}(v_i)$}{
        Compute $p_{ij}$ according to Eq.~\ref{eq:9}\;
    }
}
\For{each $v_i \in V$}{
    $P_i = \sum_{v_j \in N_G^{in}(v_i)} p_{ji}$\;
}
\While{$V \neq \emptyset$ and $|C| < M$}{
    $v^* = \arg\max_{v_i \in V} P_i$\;
    $C \leftarrow C \cup \{v^*\}$\;
    $V \leftarrow V \setminus \{\{v^*\} \cup N_G^{in}(v^*)\}$\;
    $E = E \setminus \{ (u, w), (w, u) \in E \ | \ w \in \{v^*\} \cup N_G^{in}(v^*), u \in V\}$\;
}
\For{$j = 1$ \KwTo $|C|$}{
    $\lambda_j = |\{t_i \in T |c_j = \arg\min_{c \in C} \left\|x_i - c\right\|\}|$\;
}
\Return{$C, \Lambda$}\;
\end{algorithm}

\stitle{Greedy Strategy based on Node Importance:} Based on the node importance defined in Eq.~\ref{eq:np}, we further design a greedy strategy to select the coreset, where we iteratively select the node with the largest $P_i$ value until $M$ items are found. As illustrated in Figure~\ref{fig:knng-cs}, after building a KNNG and computing the node importance for each node, we select the node $x_7$ with the largest $P_i$ value to join in $C$. Afterwards, we should not only remove $x_7$ from the graph but also all its in-neighbors $\{x_1, x_2, x_4, x_5\}$ as well as the edges related to those nodes, which are represented in gray color in Figure~\ref{fig:node-removal}. This is because the in-neighbors of $x_7$ will select $x_7$ as their representative centroid in $C$. Thus, those in-neighbors will not be considered as the candidates of $C$. 

As a summary, we present the details in Algorithm 1.  
We first initialize the coreset $C$ as an empty set in Line 1, and then compute the KNNG $G$ via the state-of-the-art method KGraph~\cite{KGraph} in Line 2. In Line 3-5, we compute each $p_{ij}$ for each edge $(v_i, v_j) \in E$ according to Eq.~\ref{eq:9}. Then, we compute the node importance $P_i$ for each node $v_i$ according to Eq.~\ref{eq:np}. In Line 8-12, we iteratively add the node $v^*$ with the largest node importance value $P_i$ to $C$ (Line 9-10), and then update the $V$ and $E$ by removing the nodes and edges related to $v^*$ and its in-neighbors (Line 11-12), until there is no node in $V$ or $M$ nodes have been selected. 
Notably, we remove the nodes in $N_G^{in}(v^*)$ because they will choose $v^*$ as their representative vector and thus will probably not appear in the coreset. Afterwards, we compute the weight $\lambda_j$ for each selected coreset member (Line 13-14). 
Finally, we return the final $C$ with at most $M$ items. To keep selection lightweight, KNNG-CS computes node importance once and performs a one-pass greedy removal. This avoids repeated benefit recomputation, which is the major bottleneck of conventional greedy methods.

\stitle{Complexity Analysis:} First, we analyze the time complexity of our method, which consists of three steps, i.e., (1) building an approximate KNNG, (2) node importance estimation for each graph node in KNNG and (3) node importance based coreset selection. According to \cite{Overlap}, the first step, i.e., buiding a KNNG, costs $O(n^{1.14}d)$. The second step costs $O(nK)$ as shown in Line 3-7. 
As to the third step, its cost is $O(nK)$, since we iteratively remove affected nodes and edges from the KNNG $G$ and there are at most $nK$ edges in $G$ to be removed. As a summary, the total cost of our method is $O(n^{1.14}d+nK)$. Notably, $K$ is pretty small in our method, which is set as 10 by default. The reason for a small $K$ value is that it will gradually reduce the number of nodes and edges removed in each iteration, so that we can select a sufficient number of coreset items. 

Second, we analyze the space complexity of our method. Three key data structures contribute to the memory occupation of our method, i.e., the KNNG $G$, the probability values $\{p_{ij}\}$ for the edges and the node importance array $\{P_i\}$. Both KNNG and the probability matrix require $O(nK)$ space, while the node importance array $O(n)$. Hence, the space complexity of our method is $O(nK)$. Since $K$ is small in our method, the space requirement is practically small as demonstrated in Section~\ref{sec:exp}.

\section{Experiments}
\label{sec:exp}
In this section, we report the results of experiments conducted on a server equipped with two Intel(R) Xeon(R) Gold 6240 CPUs @ 2.60GHz and 376GB main memory.

\stitle{Datasets:}
Four real-world datasets are used: \textbf{CovType}~\cite{covtype}, \textbf{IMDB}~\cite{leis2015good}, \textbf{Brazil}~\cite{brazil}, and \textbf{Card}~\cite{card}. 
CovType contains 581,012 samples with 54 attributes for forest cover type prediction. 
IMDB contains movie metadata for rating prediction. 
Brazil contains 98,463 orders with 9 attributes for review score prediction. 
Card contains 30,000 records with 23 attributes for payment default prediction.




\stitle{Compared Methods and Settings:}
We compare \textbf{KNNG-CS} with \textbf{Random}, \textbf{CRAIG}~\cite{jure}, and \textbf{FastCore}~\cite{cluster}. 
Random uniformly samples $M$ items; CRAIG uses partitioning only when required by memory limits; FastCore follows the authors' default settings, with candidate sample sizes of 500 for Brazil/Card and 200 for CovType/IMDB. 
For KNNG-CS, we set $M=0.01n$, $K=10$ by default, and observe that KNNG-CS is insensitive to $K$ within a reasonable range.


\stitle{Performance Metrics:}
We evaluate efficiency by coreset selection time and peak memory usage, and evaluate effectiveness by the test accuracy of a logistic regression model trained on the selected coreset, following GA-based coreset selection studies~\cite{jure,cluster}. 
We use SGD with $L_2$ regularization coefficient $10^{-5}$ and exponential learning-rate decay.  Each dataset is split into training, validation, and test sets with a ratio of $2:1:1$.


\begin{figure}[t]
    \begin{minipage}{0.24\linewidth}
        \centering
        \includegraphics[width=\linewidth]{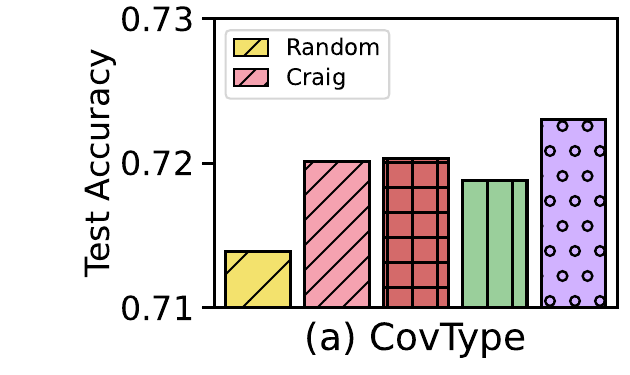}
    \end{minipage}\hfill
    \begin{minipage}{0.24\linewidth}
        \centering
        \includegraphics[width=\linewidth]{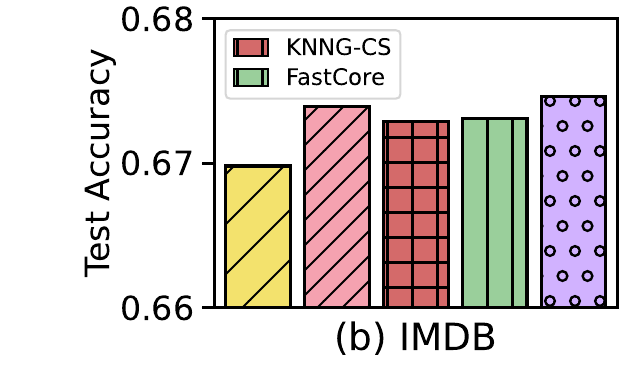}
    \end{minipage}\hfill
    \begin{minipage}{0.24\linewidth}
        \centering
        \includegraphics[width=\linewidth]{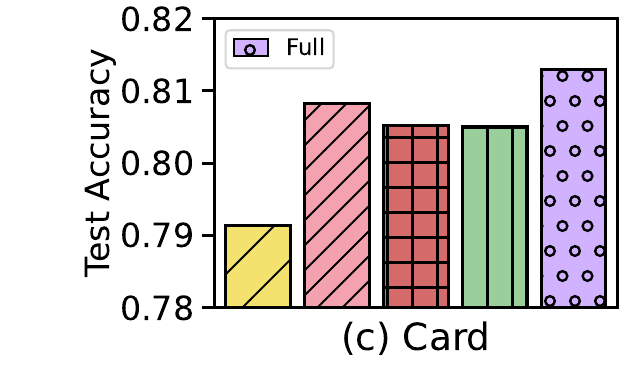}
    \end{minipage}\hfill
    \begin{minipage}{0.24\linewidth}
        \centering
        \includegraphics[width=\linewidth]{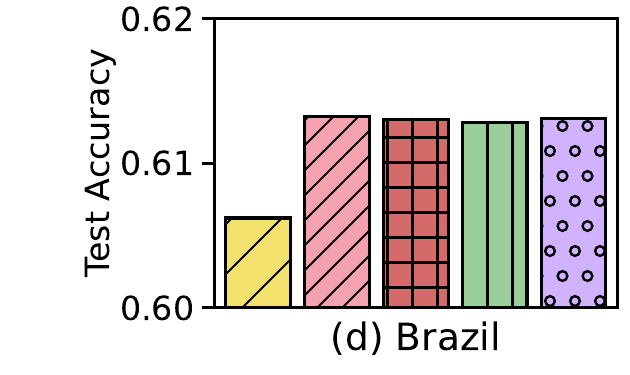}
    \end{minipage}
    \vspace{-8pt}
    \caption{Comparisons in effectiveness.}
    \vspace{-5pt}
    \label{fig:accuracy}
\end{figure}

\begin{figure}[t]
    \begin{minipage}{0.24\linewidth}
        \centering
        \includegraphics[width=\linewidth]{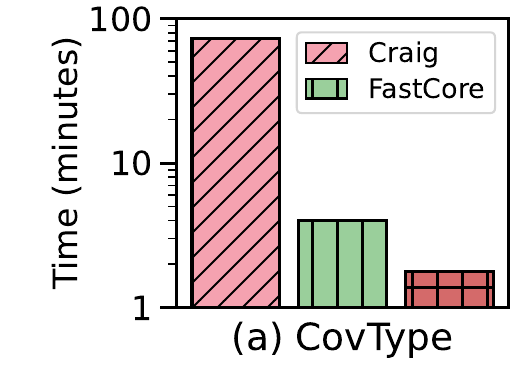}
    \end{minipage}\hfill
    \begin{minipage}{0.24\linewidth}
        \centering
        \includegraphics[width=\linewidth]{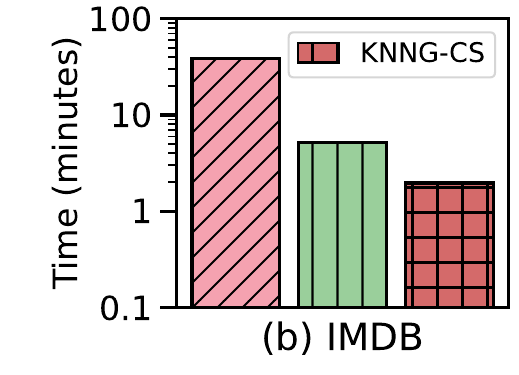}
    \end{minipage}\hfill
    \begin{minipage}{0.24\linewidth}
        \centering
        \includegraphics[width=\linewidth]{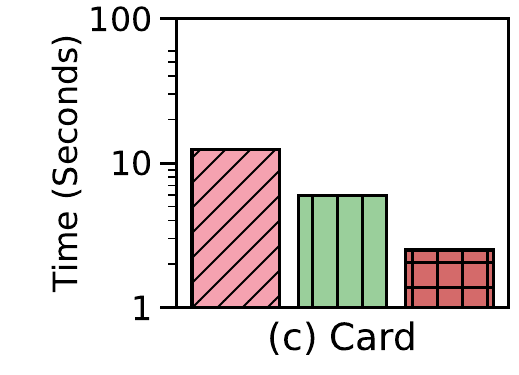}
    \end{minipage}\hfill
    \begin{minipage}{0.24\linewidth}
        \centering
        \includegraphics[width=\linewidth]{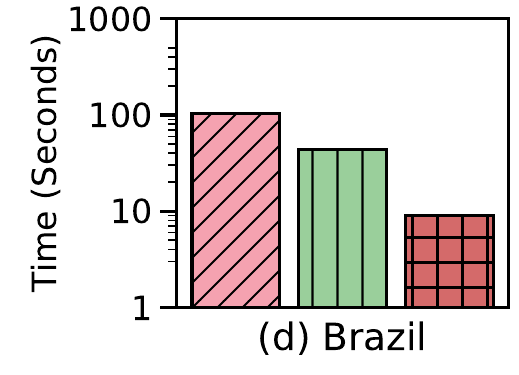}
    \end{minipage}
    \vspace{-8pt}
    \caption{Comparisons in cost consumption.}
    \vspace{-5pt}
    \label{fig:cost}
\end{figure}

\begin{figure}[t]
    \begin{minipage}{0.24\linewidth}
        \centering
        \includegraphics[width=\linewidth]{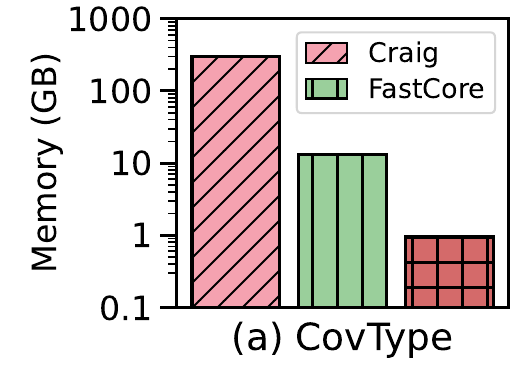}
    \end{minipage}\hfill
    \begin{minipage}{0.24\linewidth}
        \centering
        \includegraphics[width=\linewidth]{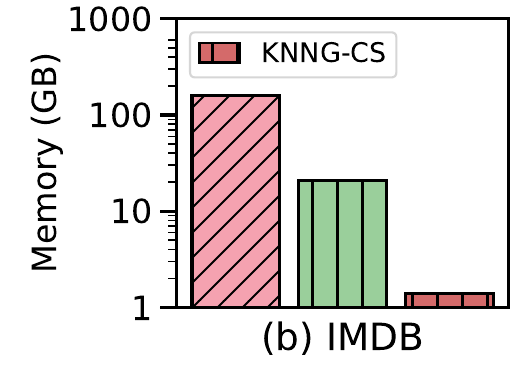}
    \end{minipage}\hfill
    \begin{minipage}{0.24\linewidth}
        \centering
        \includegraphics[width=\linewidth]{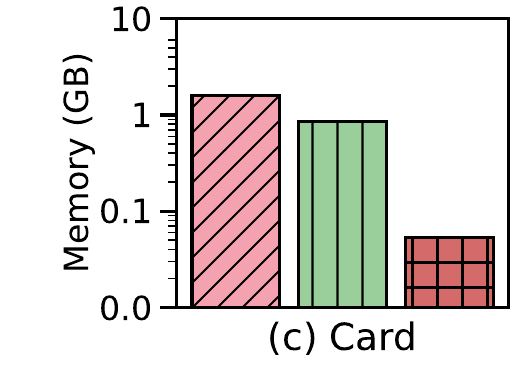}
    \end{minipage}\hfill
    \begin{minipage}{0.24\linewidth}
        \centering
        \includegraphics[width=\linewidth]{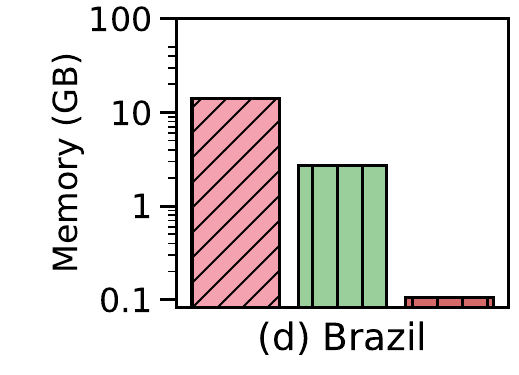}
    \end{minipage}
    \vspace{-8pt}
    \caption{Comparisons in memory usage.}
    \vspace{-3pt}
    \label{fig:memory}
\end{figure}




\stitle{Exp. 1: Overall Accuracy and Efficiency.}
We show the experimental results in the comparisons of effectiveness, cost consumption and peak memory usage in Figure~\ref{fig:accuracy}, \ref{fig:cost} and \ref{fig:memory}, respectively. First, let us discuss the comparison in effectiveness. Here, \textbf{Full} indicates no coreset selection and the model is trained on the full training set. We can see that CRAIG, FastCore and our KNNG-CS achieve comparable test accuracy with Full, due to the high quality of the selected coresets. Notably, CRAIG obtains obviously higher accuracy than FastCore and KNNG-CS on IMDB, Card and Brazil, since it takes the greedy strategy over the whole dataset with the matrix of the exact distance. FastCore achieves lower accuracy than CRAIG and KNNG-CS on CovType, Card and Brazil, since it selects each coreset item via the approximate item-cluster distance with low precision. Moreover, Random which randomly samples $M$ items, has significantly lower test accuracy than other methods.

As shown in Figure ~\ref{fig:cost}, we can see that our method outperforms CRAIG and FastCore in cost consumption. To be specific, KNNG-CS achieves 41.2x, 19.5x, 11.6x, and 5.0x speedups over CRAIG, and 2.3x, 2.6x, 4.9x, and 2.4x speedups over FastCore on the four datasets, respectively. This is because our method KNNG-CS takes a very efficient greedy strategy on top of KNNG, which could be built efficiently. On the other hand, CRAIG takes the longest execution time because it assigns an equal probability to each item as a coreset member and employs the greedy framework after computing the distance matrix, leading to very high computational complexity. 
FastCore clusters data points into a large number of clusters, i.e., $\mathcal{K}>10000$, and performs selection at the cluster level, reducing selection time and achieving 2-18x speedup over CRAIG. However, it still relies on the greedy framework, which consumes significant time to compute the upper bound matrix of item-cluster distances.

As in Figure~\ref{fig:memory}, KNNG-CS obviously requires less memory than CRAIG and FastCore. To be specific, KNNG-CS requires only 0.3\%, 0.9\%, 0.8\%, and 3.4\% peak memory of CRAIG and only 7.5\%, 6.8\%, 4.1\%, and 6.3\% peak memory of FastCore on the four datasets, respectively.
This is because CRAIG has to store the distance matrix of size up to $O(n^2)$ and FastCore has to store the item-cluster upper bound matrix of $\{u_{ik}\}$, where the number of clusters could be up to 10,000. On the other hand, KNNG-CS only needs to store a KNNG of size $O(nK)$, where the number $K$ is only 10. 

Overall, our method is clearly more efficient than its competitors, CRAIG and FastCore, in both cost and memory usage, without compromising the coreset quality. 

\stitle{Exp. 2: Effect of Coreset Size $M$.} We study the effects of the coreset size $M$ on the test accuracy of our method KNNG-CS. The results are shown in Table~\ref{tab:effcts_M}, where $M$ is represented as a function of the data size $n$. 
We can see that as the coreset size $M$ increases, the accuracy initially improves and then stabilizes, demonstrating that a small coreset can achieve sufficiently good performance due to the precision of the gradient approximation. 
Hence, we select $M=0.01 \times n$ in our experiments to balance efficiency and effectiveness.  


\begin{table}[t]
\centering
\caption{Effects of $M$ on the test accuracy}
\vspace{-5pt} 
\label{tab:effcts_M}
\begin{tabular}{|c|c|c|c|c|}
\hline
\textbf{$M$} & \textbf{CovType} & \textbf{IMDB} & \textbf{Brazil} & \textbf{Card} \\
\hline
$0.1\%n$ & 0.6964 & 0.6456 & 0.6070 & 0.7920 \\
\hline
$1\%n$ & 0.7203 & 0.6729 & 0.6130 & 0.8051 \\
\hline
$10\%n$ & 0.7218 & 0.6736 & 0.6132 & 0.8130 \\
\hline
\end{tabular}
\vspace{-2ex}
\end{table}

\begin{figure}[t]
    \centering
    \begin{minipage}{0.231\textwidth}
        \centering
        \includegraphics[width=\textwidth]{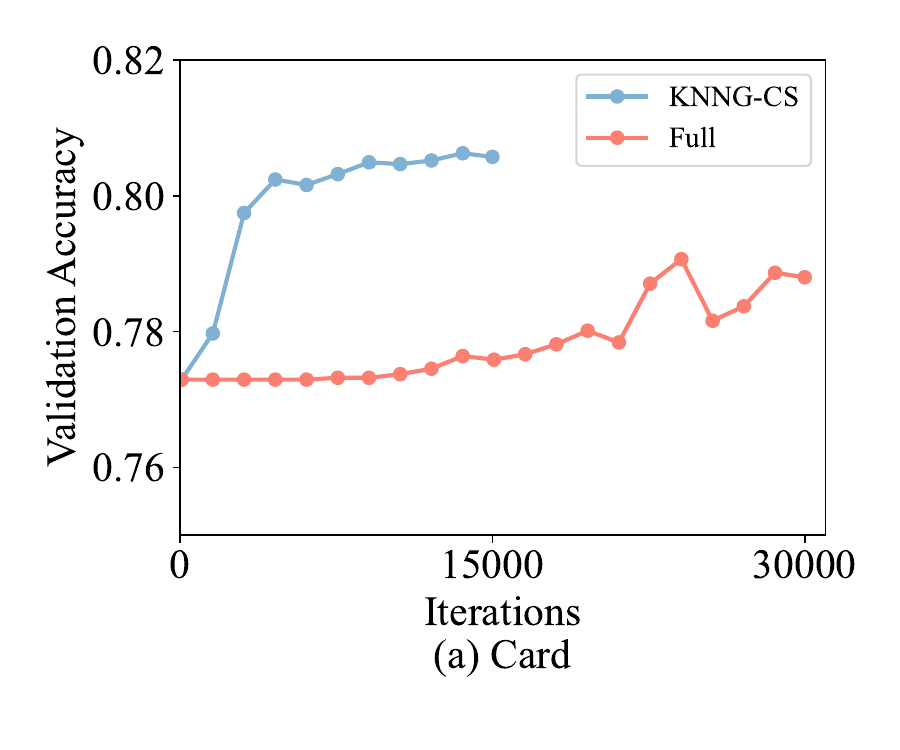}
    \end{minipage}
    \hspace{0.005\textwidth}  
    \begin{minipage}{0.231\textwidth}
        \centering
        \includegraphics[width=\textwidth]{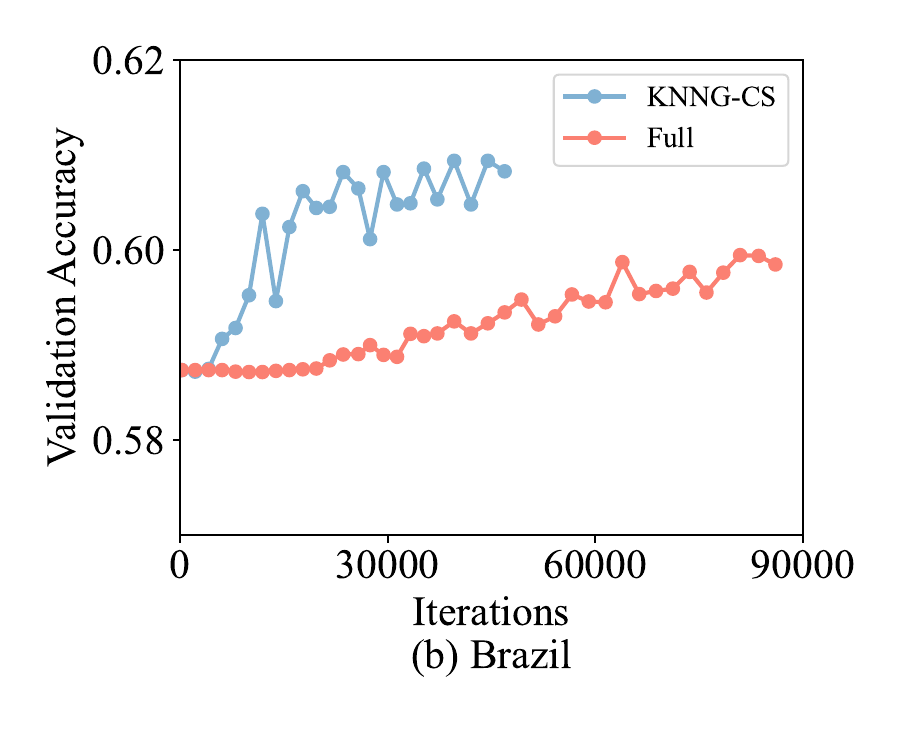}
    \end{minipage}
     \vspace{-4ex}
    \caption{Convergence Analysis}
    \label{fig:Convergence Analysis}
    \vspace{-2ex}
\end{figure}

\stitle{Exp. 3: Convergence Evaluation.} We evaluate the convergence of KNNG-CS, comparing with Full. For the Brazil and Card datasets, we set the number of epochs to 100. In Figure \ref{fig:Convergence Analysis}, the X-axis denotes the number of iterations, and the Y-axis denotes the validation accuracy. We compare the results of each method with respect to the validation accuracy, where the model achieves near-optimal test accuracy and  stabilizes in the later stages of training. We can observe that training on the coreset converges much faster than training on the full data. As Full still has to take a number of iterations to converge, we do not plot the entire process in the figure. For example, on dataset Card, KNNG-CS takes \( 1.5 \times 10^4 \) iterations to converge, but Full takes \( 1.5 \times 10^6 \) iterations. The reason is that they have the same convergence rate, so they spend a similar number of epochs converging. Since the coreset is much smaller than the full data, KNNG-CS just needs a smaller number of iterations to converge.

\section{Conclusion}
\label{sec:conclu}


In this work, we study efficient GA-based coreset selection. 
Instead of relying on dense pairwise distance matrices or large item-cluster bound matrices, we propose \textsc{KNNG-CS}, a lightweight method that exploits sparse local neighborhood structures captured by a KNNG. 
By aggregating distance-aware incoming-neighbor scores and performing one-pass redundancy-aware selection, \textsc{KNNG-CS} avoids materializing quadratic distance structures while preserving representative samples from dense local regions. 
Experiments on real-world datasets show that KNNG-CS achieves 2.3$\times$--41.2$\times$ speedups and uses only 0.3\%--7.5\% peak memory compared with state-of-the-art coreset selection methods, while maintaining comparable accuracy.

\section*{GenAI Usage Disclosure}

Large language models (LLMs) are used only for language polishing, including grammar correction, clarity improvement, and readability refinement. 
They are not used to generate technical ideas, algorithms, experiments, results, analyses, or conclusions. 
All manuscript content is carefully reviewed and verified by the authors.

\bibliographystyle{ACM-Reference-Format}
\balance
\bibliography{main}

\end{document}